\documentclass[letterpaper]{article} % DO NOT CHANGE THIS
\usepackage{aaai2026}  % DO NOT CHANGE THIS
\usepackage{times}  % DO NOT CHANGE THIS
\usepackage{helvet}  % DO NOT CHANGE THIS
\usepackage{courier}  % DO NOT CHANGE THIS
\usepackage[hyphens]{url}  % DO NOT CHANGE THIS
\usepackage{graphicx} % DO NOT CHANGE THIS
\usepackage{amssymb}
\usepackage{amsmath}
\usepackage{array}
\usepackage{booktabs}
\usepackage{xcolor}
\usepackage{array}
\usepackage{tabularx}
\usepackage{makecell}

\nocopyright

\usepackage{natbib}  % DO NOT CHANGE THIS AND DO NOT ADD ANY OPTIONS TO IT
\usepackage{caption} % DO NOT CHANGE THIS AND DO NOT ADD ANY OPTIONS TO IT
\usepackage{algorithm}
\usepackage{algorithmic}

\usepackage{newfloat}
\usepackage{listings}

\usepackage{amsthm}

\newcommand{\pro}{\textcolor{green!60!black}{\small +}}
\newcommand{\proo}{\textcolor{green!60!black}{\small ++}}
\newcommand{\prooo}{\textcolor{green!60!black}{\small +++}}
\newcommand{\con}{\textcolor{red!70!black}{\small -}}
\newcommand{\concon}{\textcolor{red!70!black}{\small - -}}
\newcommand{\conconcon}{\textcolor{red!70!black}{ - - -}}

\DeclareCaptionStyle{ruled}{labelfont=normalfont,labelsep=colon,strut=off} 
\floatstyle{ruled}
\newfloat{listing}{tb}{lst}{}
\floatname{listing}{Listing}

\title{Clustering High-dimensional Data: Balancing Abstraction and Representation\\ Tutorial at AAAI 2026}

\author {
    Claudia Plant\textsuperscript{\rm 1, \rm2},
    Lena G. M. Bauer\textsuperscript{\rm 1},
    Christian Böhm\textsuperscript{\rm 1}
}
\affiliations {
    \textsuperscript{\rm 1}Faculty of Computer Science, University of Vienna, Vienna, Austria\\
    \textsuperscript{\rm 2} Data Science @ Uni Vienna, University of Vienna, Vienna, Austria\\
    claudia.plant@univie.ac.at, lena.bauer@univie.ac.at, christian.boehm@univie.ac.at
}

\begin{document}

\maketitle

\begin{abstract}
How to find a natural grouping of a large real data set? Clustering requires a balance between abstraction and representation. To identify clusters, we need to abstract from superfluous details of individual objects. But we also need a rich representation that emphasizes the key features shared by groups of objects that distinguish them from other groups of objects.

Each clustering algorithm implements a different trade-off between abstraction and representation. Classical K-means implements a high level of abstraction - details are simply averaged out - combined with a very simple representation - all clusters are Gaussians in the original data space. We will see how approaches to subspace and deep clustering support high-dimensional and complex data by allowing richer representations. However, with increasing representational expressiveness comes the need to explicitly enforce abstraction in the objective function to ensure that the resulting method performs clustering and not just representation learning. We will see how current deep clustering methods define and enforce abstraction through centroid-based and density-based clustering losses. Balancing the conflicting goals of abstraction and representation is challenging. Ideas from  subspace clustering help by learning one latent space for the information that is relevant to clustering and another latent space to capture all other information in the data.

The tutorial ends with an outlook on future research in clustering. Future methods will more adaptively balance abstraction and representation to improve performance, energy efficiency and interpretability. By automatically finding the sweet spot between abstraction and representation, the human brain is very good at clustering and other related tasks such as single-shot learning. So, there is still much room for improvement.
\end{abstract}

 \begin{figure}[t]
  \centering
  \vspace{1mm}
    \includegraphics[width=0.7\columnwidth]{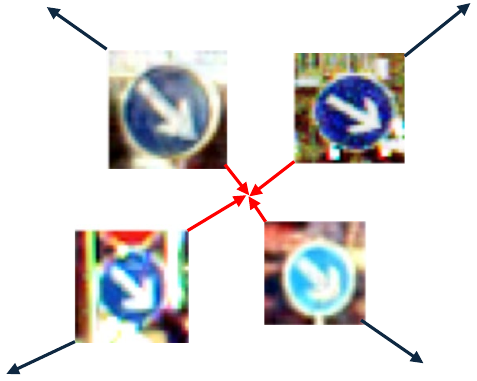}\vspace{-2mm}
   \caption{Conflicting Goals Abstraction and Representation.\vspace{-3.5mm}}
      \label{fig:conflict}
\end{figure}

\subsection*{Introduction}
 Clustering requires a balance between \emph{abstraction} and \emph{representation}. Abstraction is a central goal in natural and artificial intelligence. When confronted with massive, high-dimensional, sparse and noisy data, our brain is remarkably effective in performing abstraction, i.e. in identifying the relevant information in the flood of sensory input data. Clustering or identifying a natural grouping of massive amounts of data objects is one central approach to abstraction. To identify clusters, we need to abstract from superfluous details of individual objects, such as background or lighting in images as shown in Figure \ref{fig:conflict}. Abstraction creates attractive forces that move objects within the same cluster together (red arrows). However, to determine which variations are superfluous details and which are distinguishing features, we also need a sufficiently rich representation. Representation learning requires mapping different objects to different locations in the latent space, creating repelling forces indicated by the blue arrows.

 \begin{figure*}[t]
  \centering
  \vspace{1mm}
    \includegraphics[width=1.0\textwidth]{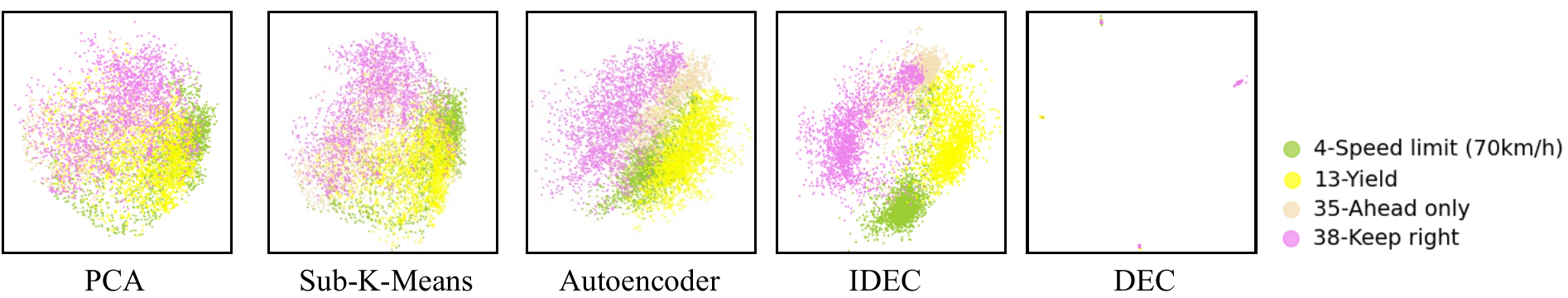}\vspace{-2mm}
   \caption{Different Representations of our Running Example Data (Subset of the German Traffic Sign Benchmark).\vspace{-3.5mm}}
      \label{fig:landscape}
\end{figure*}

 A representation containing too much information, also minor details, makes it difficult to identify the grouping. If, on the other hand, the representation is too coarse, clustering is also difficult as distinguishing features between clusters are missing. Therefore, clustering requires to find a balance between abstraction of unnecessary details and representation of key distinguishing features. As shown in Figure \ref{fig:conflict} by red arrows, we need a representation that tightly groups the traffic signs belonging to a common cluster but still preserves distinguishing differences between different objects (black arrows). Let us note that general representation learning methods and also supervised learning methods face related challenges of balancing under- and over-fitting. However, the key distinguishing feature is that we consider the challenge of balancing the level of abstraction with the accuracy of representation for the goal of clustering. As our goal is grouping of objects, the sweet spot is at a different level of abstraction/representation as for general representation learning. The goals of abstraction and representation tend to be more conflicting in the context of clustering, as abstraction is more urgently needed to find a grouping. In contrast to supervised approaches, we do not have any labels that help us identify a good trade-off.

 \begin{figure*}[t]
  \centering
  \vspace{1mm}
    \includegraphics[width=0.9\textwidth]{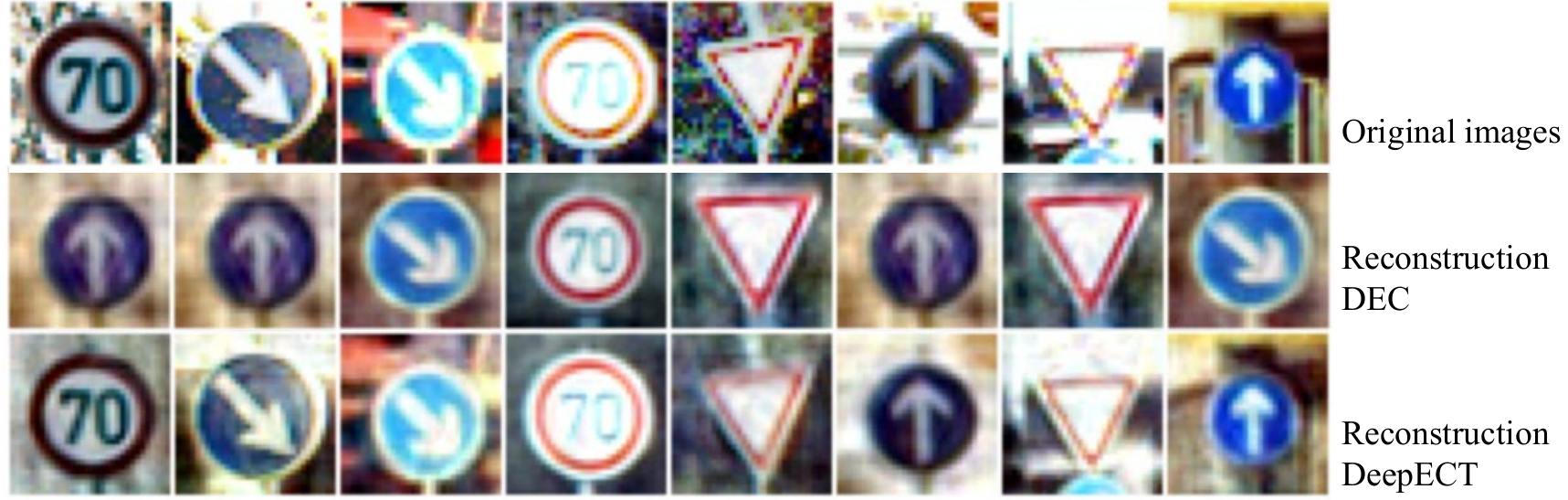}\vspace{-2mm}
   \caption{Exploring the Quality of the Learned Representation. Comparison of Image Reconstructions of Different Methods.\vspace{-3.5mm}}
      \label{fig:reconstruction}
\end{figure*}

\subsubsection*{Goal of the Tutorial}
This tutorial is intended for machine learning researchers and professionals interested in learning more about clustering high-dimensional and complex data. Clustering is a well-established research area with numerous published books and articles. Because of this abundance of literature, choosing a suitable algorithm, parametrizing it, and applying it to real data is challenging. This tutorial is intended for practitioners and researchers. Practitioners will receive an overview of different approaches to clustering high-dimensional data, along with insights into their benefits and limitations. This knowledge will enable them to select an appropriate method for their particular problem. Researchers will find starting points for contributing to the topic, as we will illustrate foundational and current approaches with Python code examples. We will summarize the evaluation methodology and provide pointers to benchmark data. We will also highlight open problems that require further research. This tutorial can therefore serve as a starting point for actively contributing to this active and fascinating research topic. 

\subsection*{Motivation: K-Means}
The tutorial starts by introducing the task of clustering, which is the task of automatically detecting a natural grouping of a set of data objects. Clustering is an unsupervised machine learning task, we do not have any labeled data. Data objects are often high-dimensional feature vectors, images or texts. Clustering of high-dimensional data is challenging due to the effects of the curse of dimensionality. Furthermore, high-dimensional data can often be clustered in multiple meaningful ways.

As a running example, we will use throughout the tutorial a subset of the German Traffic Sign Benchmark (GTSB) data set, which contains images of different traffic signs, see Figure \ref{fig:conflict} for different instances of the "keep right" sign. Each image has approximately 3,000 pixels. To review the fundamentals of clustering and demonstrate the curse of dimensionality, we will first introduce the classic K-means algorithm and apply it to this data set. Participants will be able to scan a QR code pointing to a Jupyter notebook containing the running example, \url{https://tinyurl.com/bdha8zca}. This allows them to follow the demonstration and run the algorithms in the Google Colab app, if desired. As expected, the clustering quality of K-Means is low, with an NMI of 0.28. For labeled benchmark data, the Normalized Mutual Information (NMI) of given class labels and the cluster labels assigned by the algorithm is an established quality measure for clustering results. The NMI scales between 0 and 1, where numbers close to 0 characterize poor clustering results and 1 represents a perfect clustering. For attendees who are not familiar with evaluating clustering methods on labeled benchmark data, we will briefly introduce the corresponding methodology \cite{DBLP:journals/jmlr/NguyenEB10}.

Participants will experience that successfully clustering high-dimensional data requires balancing abstraction and representation. Each clustering method has a unique trade-off between abstraction and representation. K-means clustering in the 3,000-dimensional original space involves significant abstraction because clusters are modeled by their mean vectors. However, it is unsuccessful because the original feature representation does not allow for cluster detection. Therefore, we need to learn a better representation for clustering.

\noindent One obvious option is to perform a Principle Component Analysis (PCA). Figure \ref{fig:landscape} ($1^{\text{st}}$ subfigure) shows the instances of four different classes of traffic signs after PCA dimensionality reduction to 2D space. Global PCA does not help to reveal the cluster structure as the underlying assumption that the data is one Gaussian does not apply to data consisting of multiple clusters. For the successful clustering of high-dimensional data, clustering and representation learning needs to be tightly integrated. When we know the clusters we can find a suitable representation for them; when we know a good representation, it is easy to find the clusters. Motivated by this chicken-and-egg dilemma, numerous publications focus on subspace clustering, i.e. different ways of integrating axis-parallel projections or linear dimensionality reduction methods into clustering.

\subsection{Subspace Clustering} This family of methods search for linear subspaces in high-dimensional data that contain clusters. This approach typically works well for data of moderate dimensionality up to about 100 dimensions. Some methods search for clusters in \emph{axis-parallel projections} of the original data space. Examples include the classical CLIQUE algorithm \cite{DBLP:conf/sigmod/AgrawalGGR98} and many follow-up works, e.g., \cite{DBLP:journals/isci/MaWHZZ24}. Many axis-parallel subspace clustering methods operate in a bottom-up fashion. Input parameters define a density threshold for clustering. First, the algorithm identifies dense regions in all 1D projections, and then it merges them into higher-dimensional subspaces. To efficiently explore and prune, most algorithms use the downward closure property of object density. A region that is not dense in a lower-dimensional space cannot become dense when additional dimensions are added. Axis-parallel subspace clustering methods are easy to interpret because they perform feature selection, and the original features often have meaning for domain experts. Additionally, alternative clustering is partially supported, as one data object can be assigned to different clusters in different subspaces. However, representing each cluster in its own subspace causes the relationships between clusters to be lost. These methods have limitations when applied to very high-dimensional data because the worst-case runtime complexity is exponential in the number of features. Additionally, many clusters are missed because clusters are not limited to existing in axis-parallel projections. 

Therefore, other methods search for clusters in \emph{arbitrarily oriented linear subspaces}. These methods typically integrate local PCA into clustering. For example, the 4C algorithm integrates local PCA \cite{DBLP:conf/sigmod/BohmKKZ04} into the algorithmic paradigm of the classical density-based DBSCAN \cite{DBLP:conf/kdd/EsterKSX96} algorithm. However, 4C and related approaches identify an individual subspace for each cluster, which has the drawback of not discovering relationships between different clusters.

We therefore introduced \emph{common subspace clustering} together with the algorithm Sub-K-Means \cite{DBLP:conf/kdd/MautzYPB17}. Sub-K-Means discovers a K-Means clustering together with the linear subspace that offers optimal clustering for the objective function of K-Means. This subspace is suitable for visualization and interpretation because it shows which clusters are similar to each other. Figure \ref{fig:landscape} ($2^{\text{nd}}$ subfigure from left) shows that the ground-truth classes are slightly better separated in this subspace than in that of PCA. Nevertheless, Sub-K-Means reaches its limits with this high-dimensional image dataset. The cluster quality is not actually better than that of standard K-means on this data set.

 \begin{figure}[b]
  \centering
  \vspace{1mm}
    \includegraphics[width=1.0\columnwidth]{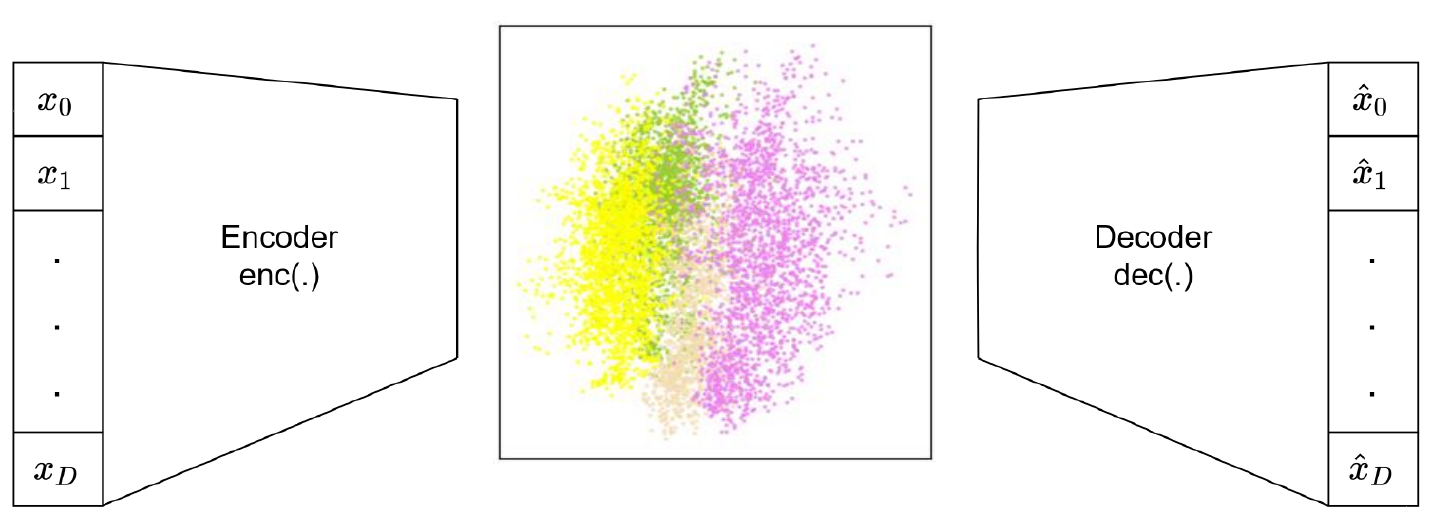}\vspace{-2mm}
   \caption{Autoencoder-based deep clustering methods integrate clustering objectives into deep autoencoders.\vspace{-3.5mm}}
      \label{fig:autoencoder}
\end{figure}

\subsection*{Autoencoder-based Deep Clustering}
For better results, deep clustering algorithms empower clustering by leveraging the representational expressiveness of deep neural networks. Deep autoencoders are multi-layer networks that learn a low-dimensional representation (typically 10 to hundred dimensions) of high-dimensional input data by minimizing the reconstruction error. An encoder network with a non-linear activation function, e.g., RELU, maps the data to the latent space. A matching decoder network reconstructs the original input object from its latent representation, see Figure \ref{fig:autoencoder}. On our running example, an autoencoder learns a representation that supports the reconstruction of images with low error. However, since the autoencoder only focuses on representation learning, there is not much cluster structure in the latent space, cf. Figure \ref{fig:landscape} ($3^{\text{rd}}$ subfigure from left). Running K-Means on the latent space of the autoencoder yields an NMI of 0.56, which is an improvement but still not sufficient. 

Deep clustering methods balance the conflicting goals of learning an expressive representation and performing abstraction simultaneously by introducing specific loss terms for clustering. These losses enforce abstraction during representation learning. In \emph{centroid-based deep clustering methods}, such as IDEC \cite{ijcai2017-243} and DEC \cite{DBLP:conf/icml/XieGF16}, the clustering loss terms encourage embedding instances of the same cluster close to their cluster centroid. These clustering losses produce representations with pronounced, well-separated clusters, see Figure \ref{fig:landscape} (right two subfigures). Centroid-based cluster losses are wide-spread in deep clustering literature as they can be easily formalized in a differentiable fashion. Unlike classical K-Means, DEC and IDEC work with soft cluster assignments to support differentiability and thus enable training of a deep neural network with mini-batches and backpropagation. The cluster loss term of DEC, IDEC and many related methods is based on the idea of cluster hardening, i.e. the latent space should have clear clusters with distinct cluster assignments. This idea is formalized by minimizing the KL-divergence between the cluster labels and a hard target cluster label distribution. After pre-training of the autoencoder, initial cluster labels are obtained by K-Means. In the following clustering phase, soft cluster labels are used and the corresponding representations are updated by hardening the cluster labels. However, the more clustered, i.e. compressed, the latent space appears, does not necessarily mean that it is actually better. In our example, DEC performs excessive abstraction. The clusters actually consist of a mix of objects with different ground-truth labels, and the NMI is only 0.58. The reason for this is that DEC only optimizes the clustering loss in the clustering phase. Although IDEC's representation looks less clustered, it performs better with an NMI of 0.60. IDEC keeps the autoencoder reconstruction error term and optimizes a weighted combination of it with the clustering loss.

Without labels, we can detect over-abstraction by evaluating the quality of the latent representations. Figure \ref{fig:reconstruction} shows some original example images (first row). Below are their respective reconstructions by DEC and DeepECT \cite{DBLP:journals/dase/MautzPB20}, a hierarchical deep clustering method. Clearly, DEC fails to reconstruct the individual images by collapsing the clusters. Each image is simply reconstructed as its cluster centroid. The individual differences between images in the same cluster are lost. DeepECT, on the other hand, achieves a good balance between abstraction and representation. Unnecessary details are omitted, yet each image retains its unique characteristics.

 \begin{figure}[t]
  \centering
  \vspace{1mm}
    \includegraphics[width=1.0\columnwidth]{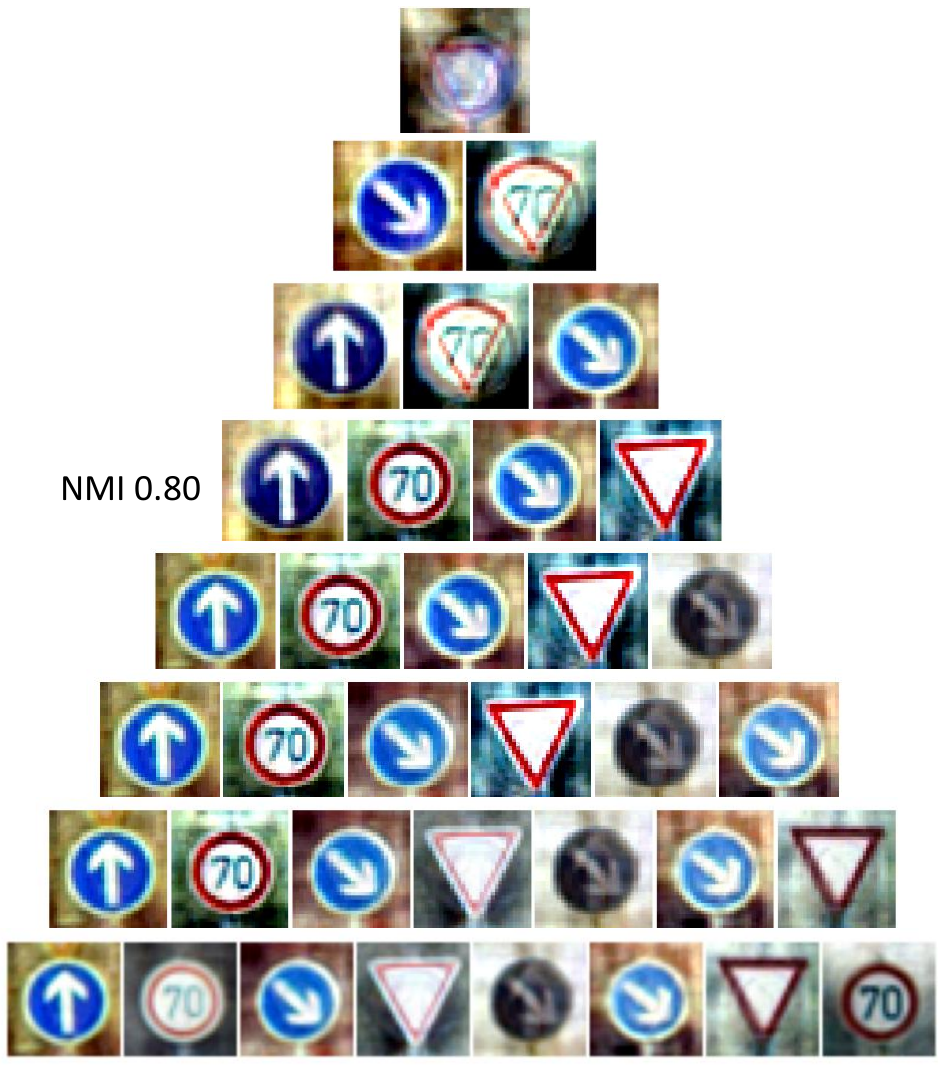}\vspace{-2mm}
   \caption{Hierarchical Cluster Tree by DeepECT. The level with 4 clusters yields an NMI of 0.80. The higher and lower levels represent meaningful clusterings as well, distinguishing signs with bright and dark backgrounds and different lighting.\vspace{-3.5mm}}
      \label{fig:clustertree}
\end{figure}

With an NMI of 0.80, DeepECT also achieves superior clustering of this data. Unlike centroid-based methods, DeepECT carefully and stepwise grows a hierarchical cluster tree in the embedded space, cf. Figure \ref{fig:clustertree} This enables the simultaneous learning of abstraction and representation. \emph{Hierarchical and density-based deep clustering methods}, see also, e.g., \cite{DBLP:conf/nips/ManduchiVRV23, DBLP:conf/pkdd/ZnalezniakRKTS23, DBLP:conf/icdm/00010MLD0P24} enable learning of hierarchical relationships between clusters. Hierarchical priors on the latent space allow for careful, local encouragement of abstraction. Therefore, these novel methods often outperform classical centroid-based deep clustering methods. 

Compared to the abundant literature on centroid-based deep clustering, only a few hierarchical, density-based or further non-parametric approaches exist. While it is relatively easy to formalize clustering losses based on attractive forces to one centroid, it is challenging to formalize density-based and hierarchical loss functions that are differentiable and trainable in a mini-batch fashion. Existing methods beyond centroid-based deep clustering are often based on limiting assumptions or parameter settings. DeepECT \cite{DBLP:journals/dase/MautzPB20} builds a hierarchy of centroid-based clusters. The DipEncoder \cite{DBLP:conf/kdd/LeiberBNPB22} and DipDECK \cite{DBLP:conf/kdd/LeiberBSBP21} integrate the Dip-test for multi-modality into autoencoders and exploit it for clustering. The Dip-test does not assume Gaussian clusters but still requires a compact unimodal cluster shape. The density-based deep clustering method SHADE \cite{DBLP:conf/icdm/00010MLD0P24}  pre-computes the cluster order in the original feature space. The hierarchical generative method \cite{DBLP:conf/nips/ManduchiVRV23} is based on a Gaussian prior (see also the section on generative deep clustering below). The contrastive method CoHiClust is based on augmentations and builds a binary tree of height defined by input parameters \cite{DBLP:conf/pkdd/ZnalezniakRKTS23} (see also the section on contrastive deep clustering below). There is a lot of room for improvement and further research. Hierarchical, density-based and non-parametric methods provide the potential for better interpretability. The latent space of these methods preserves rich information about the data structure, such as properties of clusters, local and global outliers and the relationship between clusters.

 \begin{figure*}[t]
  \centering
  \vspace{1mm}
    \includegraphics[width=0.75\textwidth]{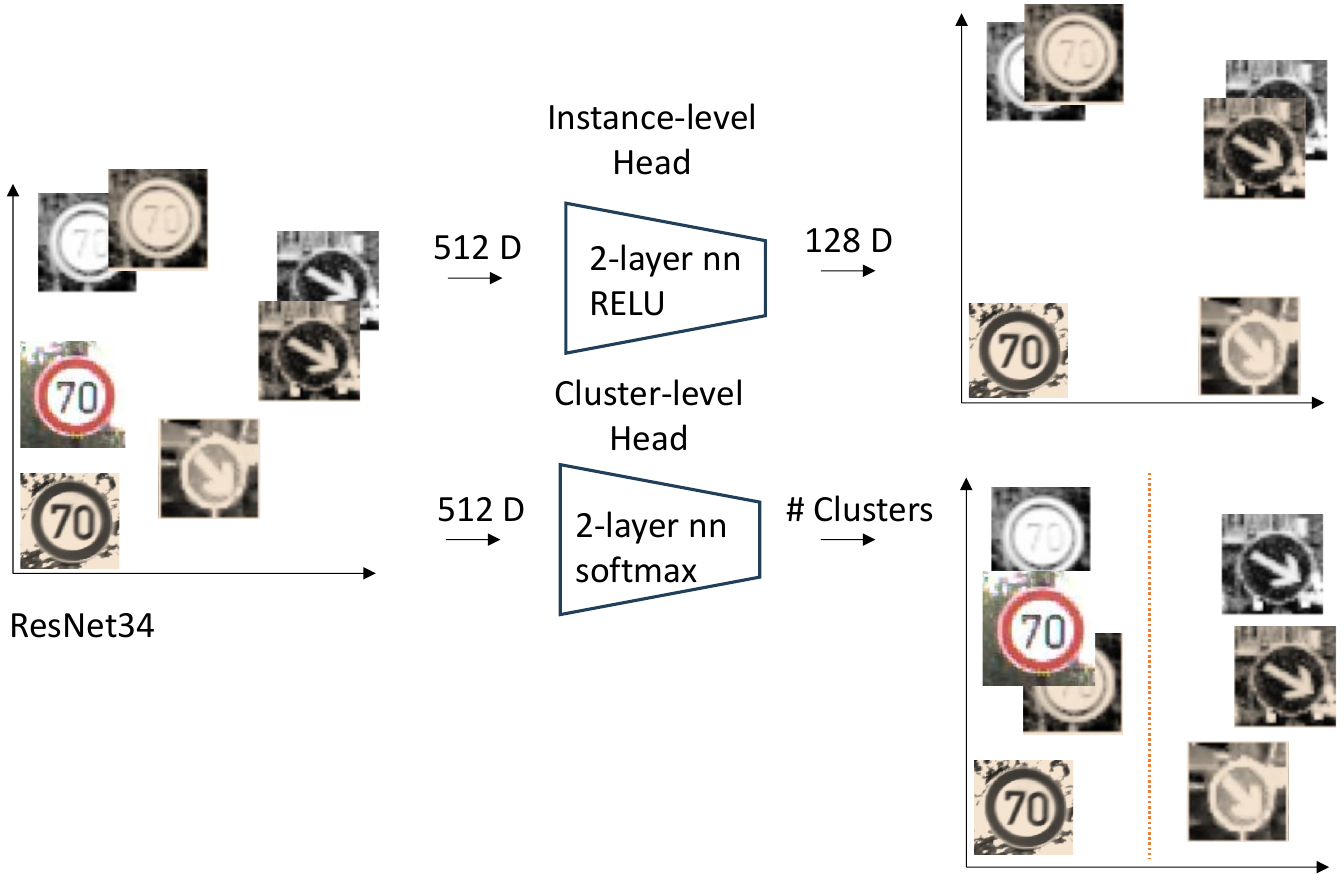}\vspace{-2mm}
   \caption{Contrastive deep clustering methods, e.g., CC integrate clustering into contrastive representation learning. In order to balance abstraction and representation, CC uses an instance-level head that is groups together the augmentations of the same image (here 2 different color augmentations), and a separate cluster level head that projects the data down to a K-dimensonal space enforcing a clear cluster structure. During test time, the cluster head outputs the cluster label of the not augmented object. \vspace{-3.5mm}}
      \label{fig:cc}
\end{figure*}

If optimizing the cluster quality is the most important goal and inspection of the latent space is not so important, ideas from reinforcement learning can boost the quality of centroid-based methods \cite{DBLP:conf/iclr/MiklautzKSLLSTP25}. The conflict between abstraction and representation introduces a difficult learning problem with changing targets. In the pre-training phase, the method only focusses on representation. In the clustering phase, the focus is mostly on abstraction. From reinforcement learning it is well known that neural networks suffer from dying neurons and lacking flexibility for further learning when targets change. Weight resets help to maintain neural plasticity. Our method BRB performs periodic weight resets of the penultimate layer and can be integrated into centroid-based autoencoder and contrastive deep clustering methods. BRB prevents over-commitment by increasing the intra-cluster variation while preserving cluster separation. It offers improvement of cluster quality and allows training from scratch at the expense of 1\% of runtime overhead.

Another interesting research direction towards avoiding the conflict between abstraction and representation is learning multiple representations. Our framework ACe/DeC \cite{DBLP:conf/ijcai/MiklautzBMTBP21} can be combined with autoencoder centroid-based deep clustering methods. It splits the latent space into two spaces. The clustered space for capturing the information relevant for clustering and the shared space for all further variations in the data. Experiments show that this explicit disentanglement of information results in superior cluster performance and allows isolating the information relevant for clustering in very low-dimensional spaces. The method also becomes more robust against hyperparameter settings and models train faster. A natural extension of this idea is learning multiple alternative clusterings with associated representations. We have developed ENRC as foundational algorithms in this area \cite{DBLP:conf/aaai/MiklautzMABP20} and applied them to the challenge of creating a classification system for medieval glass beads together with partners from archeology \cite{DBLP:conf/dsaa/MiklautzSLTSWRBP23}.

\begin{figure*}[t]
  \centering
  \vspace{1mm}
    \includegraphics[width=0.75\textwidth]{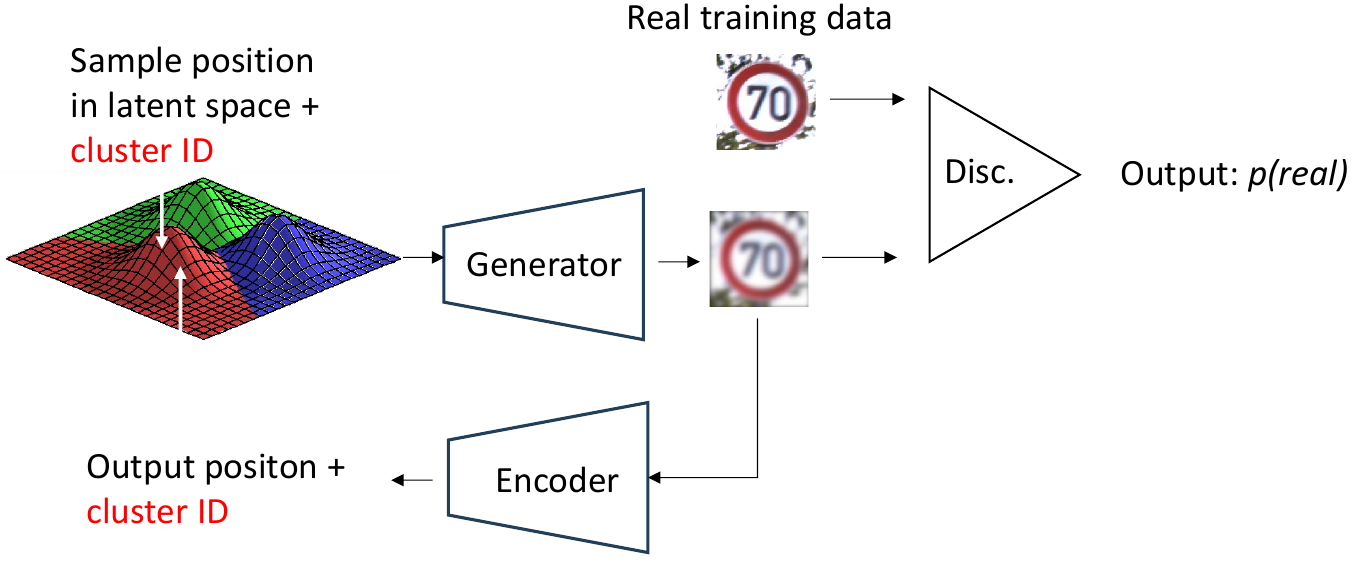}\vspace{-2mm}
   \caption{ClusterGAN extends the architecture of a generative adversarial network by an encoder network that enforces a clear cluster separation.\vspace{-3.5mm}}
      \label{fig:generative}
\end{figure*}

\subsection*{Contrastive Deep Clustering}
Contrastive deep clustering methods are powerful alternatives to autoencoder-based methods for data types where semantically meaningful augmentation methods exist, especially natural images \cite{DBLP:conf/eccv/GansbekeVGPG20, Li_Hu_Liu_Peng_Zhou_Peng_2021, DBLP:journals/pr/CaiGZF26, Dang_2021_CVPR}. Contrastive representation learning methods take positive and negative pairs of data objects as input to learn a representation where positive pairs representing similar objects are close to each other and negative pairs representing dissimilar objects are mapped far apart from one another. For supervised learning where labels of training data objects are available, pairs can be easily constructed exploiting the label information. As labeled data is not available, contrastive deep clustering methods usually construct pairs based on augmentations. The method CC \cite{Li_Hu_Liu_Peng_Zhou_Peng_2021}, for instance, works with two different augmentations that serve as input to a Siamese network architecture using an encoder with shared weights. The encoder is implemented as a ResNet34 that outputs a 512-dimensional feature vector for each image. Positive pairs are augmentations of the same image, negative pairs are augmentations of different images. Most contrastive deep clustering methods are based on similar architectures that are also common in contrastive representation learning methods, e.g. SimCLR \cite{10.5555/3524938.3525087}, MoCo or BYOL. A representation trained on the task of instance discrimination needs more abstraction to be suitable for clustering.

The contrastive deep clustering methods differ in their approaches how to learn cluster-friendly representations from this basis. CC adds two separate neural networks operating on this contrastive representation (see Figure \ref{fig:cc}). The Instance-level Head is a 2-layer network that ensures instance discrimination by projecting the 512-dimensional vector to 128 dimensions based on an Information-Noise-Contrastive Estimation loss as commonly employed in contrastive representation learning. The Cluster-level Head enforces a much stronger abstraction by projecting the 512-dimensional vectors to a low-dimensional space having the dimensionality set as the number of clusters, which is an input parameter in CC. The Cluster-level Head is also implemented as a two-layer network with a softmax activation function. The resulting $K$-dimensional feature vectors can therefore be interpreted as soft cluster assignments. During training, this network is optimized with an objective function encouraging the consistency of cluster assignments of the objects in both augmentations and rewarding a high entropy of the cluster assignments, to avoid that every object is assigned to the same cluster. During test time, test objects are just passed through the encoder. The Cluster-level Head outputs the cluster label. Other methods completely drop the instance-level contrastive loss after pre-training in order to solely focus on clustering. SCAN \cite{DBLP:conf/eccv/GansbekeVGPG20} and MNN \cite{Dang_2021_CVPR} are based on the idea of mining nearest neighbors. The representation is updated such that the cluster assignments of objects and their neighbors are consistent. The recent method discDC \cite{DBLP:journals/pr/CaiGZF26} performs linear discriminant analysis within an encoder/decoder framework in order to enhance the cluster separation.

\subsection*{Generative Deep Clustering}
Generative deep clustering methods assume a prior on the data distribution of the clusters in the latent space, typically a Gaussian mixture Model. This assumption supports learning the latent space representation of the data and the cluster labels simultaneously. The foundational methods VaDE \cite{DBLP:conf/ijcai/JiangZTTZ17} and ClusterGAN \cite{DBLP:conf/aaai/MukherjeeALK19} represent two different approaches to learning representations and cluster assignments. VaDE extends a variational autoencoder to support clustering by assuming a Gaussian mixture model with K components. The assumed generative process of the data is to first choose a cluster identifier, then choose a latent representation that consists of the parameters (mean and variance) of the chosen cluster and then sample a data object from this distribution.  Variational autoencoders (VAEs) have similar encoder-decoder architectures as conventional autoencoders. The key difference is that the encoder maps each input object to a probability distribution, typically Gaussian, i.e. it outputs the mean and the standard deviation of the latent variables representing the input object. For clustering, VaDE assumes a mixture of Gaussians with hard cluster assignment. The encoder additionally outputs the cluster label representing the assignment to one out of K Gaussians. The objective function of VaDE consists of a reconstruction term that maximizes the log-likelihood of the original data given the latent representation and a regularization term that minimizes the KL-divergence of the latent representation from the Gaussian prior. The reconstruction term is not straightforward differentiable w.r.t. the parameters of the Gaussian that the encoder needs to learn because learning would require sampling from the probability distribution. Inherited from VAEs, VaDE exploits the reparameterization trick to enable differentiability of the reconstruction term. This trick introduces a separate Gaussian noise variable that does not depend on parameters learned by the encoder. Thereby stochastic sampling is turned into a differentiable transformation that can be learned by backpropagation. Nevertheless, stable learning of the generative model is often challenging and requires much training data to prevent overfitting. For clustering multi-view data, VAE-based methods have recently attracted much attention, as the probabilistic generative model supports identifying end exploiting common information across views, e.g., \cite{DBLP:conf/iccv/Xu0TP0Z021, DBLP:conf/aaai/XuW0HLF024}.

As an alternative to variational autoencoders, another line of generative deep clustering methods are based on generative adversarial networks (GANs). A GAN consists of two neural networks that compete with each other. The generator takes as input a probability distribution and generates data objects with the goal to fool the discriminator. The discriminator tries to distinguish the fake data objects generated by the generator from real data objects. To construct a GAN architecture that support clustering, ClusterGAN \cite{DBLP:conf/aaai/MukherjeeALK19} introduces a mixture of discrete and continuous variables in the latent space: the discrete cluster ID and the continuous parameters of the corresponding Gaussian. Similar to VaDE, the latent space is modeled as a mixture of Gaussians with K components. ClusterGAN adds to the GAN generator-discriminator architecture an encoder network which receives the data objects generated by the generator and outputs their latent code consisting of the cluster ID and the parameters (see Figure \ref{fig:generative}). The encoder enforces the separation of clusters as they need to be non-overlapping in order to be reliably predictable from the generated objects. The loss function of ClusterGAN sums up the three losses of the generator, the discriminator and the encoder with hyperparameters representing the weighting - and also specifying a balance between representation and abstraction. High weighting for the encoder enforces clearer cluster separation at the expense of generating realistic data samples. Recently, a hierarchical deep clustering algorithm based on multiple generators has been proposed \cite{DBLP:conf/aaai/MelloAM22}.

\subsection*{Related Tutorials and Surveys}
 We previously gave a tutorial on the application of deep clustering algorithms with ClustPy at the CIKM 2023 conference \cite{DBLP:conf/cikm/LeiberMPB23}. The CIKM 2023 tutorial focused on autoencoder-based deep clustering algorithms. We recently published an updated survey on autoencoder-based deep clustering algorithms on ArXiv \cite{DBLP:journals/corr/abs-2504-02087}. In this tutorial, we provide a broader overview of the challenges of clustering high-dimensional data and current solutions. 

For further reading, we recommend survey papers that cover parts of the tutorial. There are several surveys on subspace clustering, and related topics including feature selection for clustering and bi-clustering, e.g., \cite{DBLP:journals/sigkdd/ParsonsHL04, DBLP:journals/tkdd/KriegelKZ09} with a general focus and \cite{DBLP:journals/tcbb/MadeiraO04, DBLP:journals/widm/DerntlP16} with focus on applications from biology and neuroscience, and \cite{DBLP:journals/csr/YuRWDZ24} with focus on discovering multiple alternative clustering solutions. Deep clustering is a very active research area. Surveys like \cite{DBLP:journals/csur/ZhouXZCLBWWZE25, DBLP:journals/ijon/WeiZHZ24, DBLP:journals/inffus/ChowdhuryGD25} summarize different basic approaches and architectures. While \cite{DBLP:journals/csur/ZhouXZCLBWWZE25, DBLP:journals/ijon/WeiZHZ24} focus on general deep clustering, \cite{DBLP:journals/inffus/ChowdhuryGD25} specifically consider approaches for multi-view data.

\subsection*{Summary and Outlook}
 Starting from classical K-Means, we have experienced the benefits and drawbacks of subspace and deep clustering methods. Table \ref{tab: characteristics} provides a summary. Classical clustering methods like K-Means or DBSCAN break down in effectiveness on high-dimensional data such as images. However, these classical methods also offer attractive benefits. Their result is easy to interpret as the clustering is performed in the original feature space without any transformation. The carbon footprint is much smaller than that of more advanced methods. Often, clustering is possible on a single CPU without parallelization, and the parameterization of classical methods is also relatively easy as rather few parameters need to be set. Also, the topic of automatic parameterization in classical clustering is well explored and solutions based on rules of thumb, Bayesian statistics or Minimum Description Length often give satisfactory results.  

 Subspace clustering methods enable clustering of data up to about 100. Searching for clusters in linear subspaces, the interpretability of the results is still relatively easy and they are still quite runtime efficient without the need for GPUs. Deep clustering methods support high-dimensional data with thousands of dimensions at the expense of lower interpretability due to non-linear feature transformations, much higher carbon footprint and difficult parameterization.
 
 To combine the benefits of the different categories of methods, in the medium term, we envision hybrid methods that integrate efficient and interpretable subspace clustering and expressive deep clustering. Existing methods implement a pre-defined and fixed trade-off between abstraction and representation - starting from classical clustering methods which only focus on abstraction without any representation learning, over linear subspace, to non-linear representations in deep clustering. Hybrid methods will need to find the best trade-off between abstraction and representation in a flexible way during runtime. Also, this trade-off will not be global but specific and local; not only specific for a data set but also for certain clusters and even for individual data objects.
\begin{table}[t]
\centering
\footnotesize
\setlength{\tabcolsep}{3pt}
\renewcommand{\arraystretch}{1.12}

% Row-label column made narrower; remaining columns auto-share the rest
\begin{tabularx}{\columnwidth}{>{\raggedright\arraybackslash}p{0.22\columnwidth} *{4}{>{\centering\arraybackslash}X}}
\toprule

&{High-}&{Inter-}&{Carbon}&{Para-}\\
&{dimensional}&{pretability}&{Footprint}&{meterization}\\
\midrule
{classical} & \conconcon & \prooo & \prooo & \con \\
{subspace}  & \pro    & \proo  & \proo  & \concon \\
{deep}      & \prooo  & \pro   & \conconcon& \conconcon \\
{hybrid}    & \prooo   & \prooo & \pro   & \concon \\
\bottomrule
\end{tabularx}
\caption{Summarized Characteristics of classical, subspace and deep clustering methods motivating the need for research on novel hybrid methods.}
\label{tab: characteristics}
\end{table}

\bibliography{sample-bibliography}

\end{document}